\documentclass[runningheads]{llncs}
\usepackage{graphicx}
\usepackage{amsmath,amssymb} 
\usepackage{color}
\usepackage[width=122mm,left=12mm,paperwidth=146mm,height=193mm,top=12mm,paperheight=217mm]{geometry}

\newcommand{\vI}{I}
\newcommand{\vx}{x}
\newcommand{\vF}{{\bf F}}
\newcommand{\mF}{\Omega_f}
\newcommand{\mH}{\Omega_h}
\newcommand{\mB}{\Omega_b}
\newcommand{\vr}{\rho}
\newcommand{\vsh}{s}
\newcommand{\vn}{n}
\newcommand{\vl}{\theta}

\newcommand{\renderSH}{\mathrm{SH}}
\newcommand{\render}{\mathrm{render}}

\newcommand{\comment}[1]{}

\newcommand{\boldstart}[1]{\vspace{0.1in} \noindent \textbf{#1}}

\newcommand*{\affaddr}[1]{#1} 
\newcommand*{\affmark}[1][*]{\textsuperscript{#1}}

\begin{document}
\pagestyle{headings}
\mainmatter

\title{A Visual Representation for Editing Face Images} 

\titlerunning{A Visual Representation for Editing Face Images}

\authorrunning{Jiajun Lu, Kalyan Sunkavalli, Nathan Carr, Sunil Hadap, David Forsyth}



\author{Jiajun Lu\affmark[1], Kalyan Sunkavalli\affmark[2], Nathan Carr\affmark[2], Sunil Hadap\affmark[2], David Forsyth\affmark[1]}

\institute{\affaddr{\affmark[1]University of Illinois at Urbana Champaign}, \affaddr{\affmark[2]Adobe Research}\\
    \email{\affmark[1]\{jlu23, daf\}@illinois.edu}, \email{\affmark[2]\{sunkaval, ncarr, hadap\}@adobe.com}\\
}

\maketitle

\begin{abstract}

We propose a new approach for editing face images, which enables numerous exciting applications 
including face relighting, makeup transfer and face detail editing. Our face edits are based on a 
visual representation, which includes geometry, face segmentation, albedo, illumination and detail map.
To recover our visual representation, we start by estimating geometry using a morphable face model, 
then decompose the face image to recover the albedo, and then shade the geometry with the albedo and 
illumination. The residual between our shaded geometry and the input image produces our detail map, 
which carries high frequency information that is either insufficiently or incorrectly captured by 
our shading process. By manipulating the detail map, we can edit face images with 
reality and identity preserved. Our representation allows various applications. First, it allows a user to 
directly manipulate various illumination. Second, it allows non-parametric makeup transfer 
with input face's distinctive identity features preserved. Third, it allows non-parametric modifications to the 
face appearance by transferring details. For face relighting and detail editing, we evaluate via a user
study and our method outperforms other methods. For makeup transfer, we evaluate via an online attractiveness evaluation
system, and can reliably make people look younger and more attractive. We also show extensive 
qualitative comparisons to existing methods, and have significant improvements over previous techniques.

\end{abstract}

\section{Introduction}

Post-capture editing of lighting, makeup and face appearance is an important component of photographic retouching, visual effects, and image compositing. However, it is a very 
challenging problem because realistic editing usually requires precise estimates of scene geometry, material properties, and lighting. This is especially true for faces where even 
small artifacts lead to perceptually implausible results. 
\begin{figure}[t]
\centering
   \includegraphics[width=0.95\textwidth]{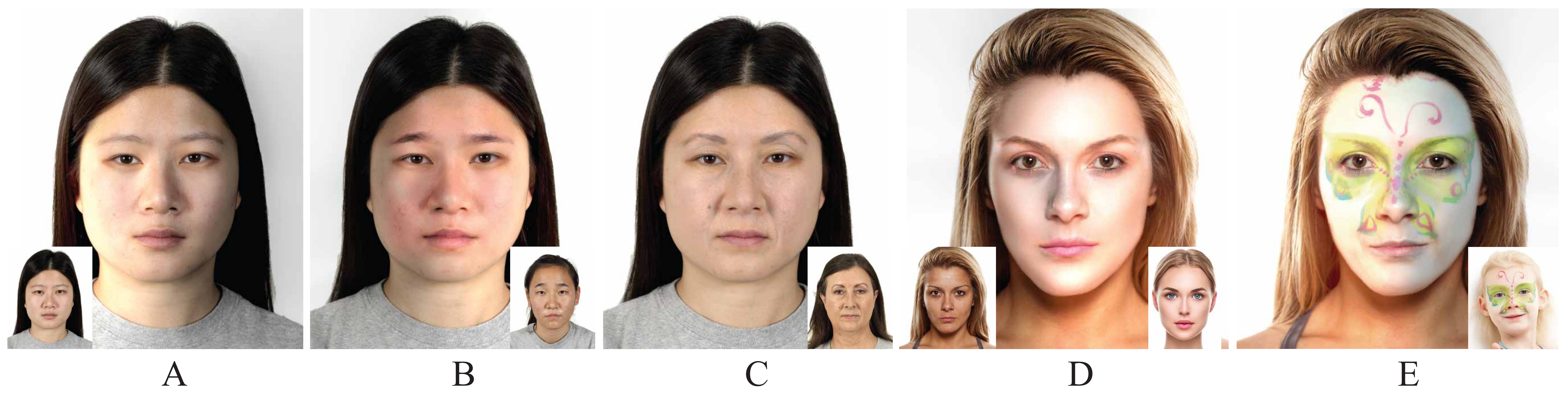}
\vspace{-3ex}
   \caption{Our paper proposes a visual representation to automatically relight a single face image with face details preserved (A), edit face detail by transferring detail maps from a 
different subject (B,C), and transfer face makeup from a reference (lower right inset) image to an input (lower left inset) image (D,E). 
Our method keeps the image's realism during all three operations, and preserves the input face's identity (detail around nose, eyes and mouth) while transferring makeup.  {\em Best viewed at high resolution, in color.} }
\label{fig_teaser}
\vspace{-3ex}
\end{figure}

Our representation exposes natural ``hooks'' for image editing. We align a morphable face model to the face in an image, then 
recover albedo and shading maps for the image, which are used in conjunction with the registered face model to estimate scene illumination. 
We render the appearance of the face given the estimated geometry, lighting, and albedo using Lambertian shading. 
However, current technology does not allow recovering a face geometry up to the scale of wrinkles, pores, etc. These small details are
important parts of what make a face individual (for exmaple, careful preservation of detail around eyes, nose and mouth ensures that changing
face makeup by editing albedo does not change the input's identity). Thus we compute a detail map
containing image information not represented by recovered albedo, shading and geometry. The idea behind our method is that editing operations 
enabled by such a detail map can be made reasonably robust against the accuracy of the models used to produce the detail map. 
We demonstrate three types of face editing tasks: relighting, makeup transfer(we take 
makeup from a reference and place it on an input) and detail editing (we change surface detail on a face), see Fig.~\ref{fig_teaser}.

We evaluate our method by: qualitative evaluations of result images; qualitative comparisons to the results of existing methods; 
quantitative results of a user study, where we compare the ability of persons to distinguish between edited faces and untouched images; 
quantitative results of an online system to evaluate the effect of makeup on age and attractiveness. 
These experiments show that our technique is able to produce results that have significant improvements over previous techniques in terms of perceptual realism.
For relighting and detail transfer, our results are hardest to distinguish. For makeup transfer, we can reasonably make people look younger and more attractive. 

\boldstart{Contributions:}

\begin{enumerate}
\item We propose an effective visual representation for face (with face detail maps related to face realism and face identity) that can realistically and reliably perform multiple tasks
automatically, including face relighting, makeup transfer and face detail editing. Previous methods are usually effective only on a single task. 

\item Our model is the first model that allows  interactive relighting of a whole single face image. We propose the first visual quality based quantitative 
way to evaluate face relighting results, and our method produces state of the art face relighting results.

\item Our makeup transfer preserves the person's identity and creates realistic results with various makeup images ( such as normal makeup, unusual makeup, etc), 
which previous methods have problems dealing with. Even for normal makeup, we create state of the art results. 

\item Our model also enables us to edit face details to achieve various effects, such as making the person look younger or older, changing skin properties and fine-scale geometric 
details. By grouping and choosing appropriate source images, we can control tons of editing operations to create desired images. Compared to previous method,
we support a larger variety of tasks, and perform as well or better. 
\end{enumerate}

%
\section{Background}

\boldstart{Face relighting:} Ratio image~\cite{M+G}~\cite{R+S} is a standard method to relight face images.
Numerous variants are available to: improve performance under extreme lighting conditions~\cite{Wang3};  sew together local ratio estimates to resolve 
difficulties created by small errors in registration~\cite{Chen}; relax the requirement that reference images are registered~\cite{Sto};
allow expression transfer~\cite{Liu}; relight temporal sequences, using a large database of reference images~\cite{peers}.

The requirement for two reference face images can be relaxed in ratio image methods. Wen et al.~\cite{wen} describe a method that replaces the reference face images with 
reference spheres. Chen et al.~\cite{Chen} describe a method that requires one reference face, in the target illumination.
Li et al.~\cite{Li} estimate low spatial frequency components using logarithmic total variation. Wang et al.~\cite{Wang2} relight single images by recovering face geometry. 
Each method is demonstrated only on masked faces. Chai et al.~\cite{Chai15:Hair} focus on accurate geometric reconstruction 
of face and hair, which is used to relight portraits. In contrast to our method, there is no explicit detail map representation that captures rendering errors for the inferred model.

For the methods described, evaluation is by qualitative comparison, and (for~\cite{Wang2}~\cite{Wang3}) by quantitative studies of face recognition results (the 
percentage recognition accuracy improved by relighting).  In contrast, our quantitative evaluation tests whether people are fooled into thinking that relit faces are real images.

\boldstart{Makeup transfer:} Analyzing and modifying images to improve perceived beauty is an active area (review in~\cite{liu2014fashion}).
A variety of methods allow interactively applying makeup to the input face. By recovering and then editing multiple reflectance layers, 
makeup can be successfully simulated~\cite{Li15:Makeup}. TAAZ allows users to apply chosen makeup components to face images
interactively (\url{www.taaz.com}). In contrast, our method allows automatic transfer of makeup from reference to input image.

Automated makeup transfer methods typically use 2D face models.  Tong et al.~\cite{tong2007example} transfer makeup using a ratio image.
Guo et al.~\cite{guo2009digital} transfer makeup by decomposing images into several layers, then blend layers and recomposite. 
Their detail layer is obtained by image filtering. Liu et al.~\cite{liu2014wow} register facial components independently, and synthesize made-up faces using database. 
Liang et al.~\cite{liang2014facial} enhance facial skin by suppressing a detail layer, estimated using image filtering. 
Shih et al.~\cite{Shih14:StyleTransfer} adjust face image subband energies to match the makeup. 
Scherbaum et al.~\cite{scherbaum2011computer} suggest makeup for a 3D morphable face model based on a collection of 3D example morphable face models. 
In contrast, our detail layer is obtained from rendering residual, and contains material and identity information; we only need 2D images and allow relighting;
we offer various makeup types and larger number of evaluation images.

\boldstart{Appearance editing:} Face expression editing is possible using ratio images, and produces good results~\cite{Liu}.
Bitouk et al. achieve expression and illumination transfer by replacing a face image with a similar image (with appropriate expression and illumination parameters) from an
enormous collection~\cite{Bitouk}. Yang et al. warp local face components from other images to transfer expression~\cite{Yang}. 
Kemelmacher-Shlizerman et al. combine models of illumination subspaces of faces at various ages~\cite{Kemelmacher14:Age}. 
In contrast, our editing is achieved by warping and applying detail maps. 

\boldstart{Detail maps:} Fine relief on surfaces creates small shadows that appear to be strong material cues to human viewers (eg.~\cite{Pont}).
Representations of this detail can be used to slightly improve albedo estimates and material classifiers~\cite{zichengshadingdetail}.
Liao et al.~\cite{Liao} demonstrated that residual map can help relight some objects, because convincing 
material appearance is more important to human subjects than physical accurate shading fields.
Boyadzhiev et al.~\cite{Boyadzhiev2015} edit surface material (e.g. skin oilness or blemishes on faces) by adjusting selected image bands typically at high 
spatial frequencies. In contrast, for relighting we preserve those bands and adjust low spatial frequencies; for makeup and detail transfer we use a 
non-parametric model of those bands.

%
\section{Recovering a Visual Representation}

We need to recover our visual representation of faces to build our various face editing pipelines.
Let $F$ define the aligned 3D face geometry, which allows us to build coherent face coordinates, and estimate face normals. 
Let $\vsh(\vx)$ and $\rho(\vx)$ define image shading and albedo, which are recovered by some intrinsic image process. 
Let $\theta$ represent the image illumination, which can include luminares such as spherical lighting and point light source.
Finally, we define a detail map $\vI_d(\vx)$, which carries identity and finer scale geometry.
We render a representation $(\rho, F, \theta, \vI_d)$ to produce a new image by the rule
\begin{equation}
\vI(\vx) = \render(\rho(\vx), F, \theta) + \vI_d(\vx)
\label{eq_detail1}
\end{equation}
where $\render()$ denotes any reasonable modern 3D rendering process. 
Next, we will talk about recovering each component in detail, and example components can be found in makeup transfer pipeline in Fig.~\ref{figmaps}.

\begin{figure}[t]
\centering
\includegraphics[width=1.0 \textwidth]{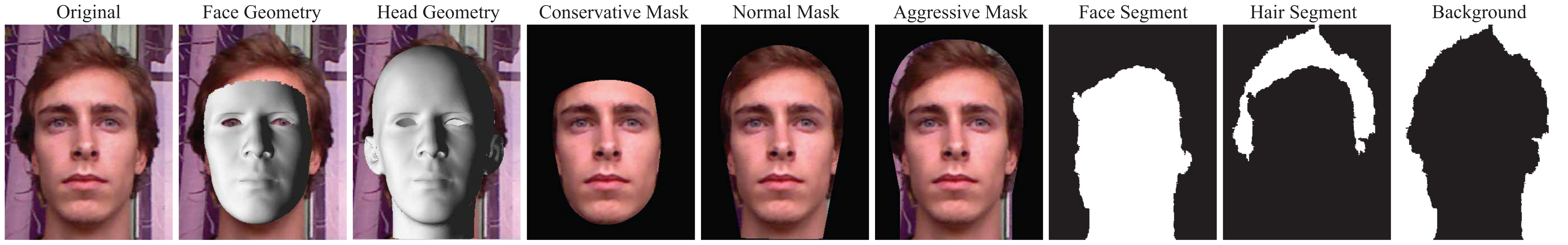}
\caption{The reconstructed face geometry registered to images, and face masks in face segmentation. }
\label{fig_geometry}
\vspace{-5ex}
\end{figure}

\boldstart{Face geometry:}  We reconstruct the geometry of the face in the input image using a morphable face
model~\cite{Blanz1999Morphable} built from the FaceWarehouse data~\cite{Cao14:FaceWarehouse}.
 The model produces a 3D face mesh $\vF(\omega_i, \omega_e)$ that is a function of identity parameters, $\omega_i$, and 
expression parameters, $\omega_e$.  FaceTracker~\cite{FaceTracker} is used to detect landmarks $l_i$ on the face, 
then we recover parameters and pose of the face mesh by minimizing the distance between the projected landmark vertices and their 
corresponding landmark locations on the image plane.  We propose a Gaussian Mixture Model as a regularizer. The GMM fits to the 
values of these parameters for the FaceWarehouse data, and a scaled negative probability density value is used as our regularizer.
Our morphable face model gives us vertex-to-vertex correspondences between all face meshes. By projecting image pixels to the 
face meshes (via barycentric coordinates), we get pixel-to-pixel correspondence between all images, see Fig.~\ref{fig_geometry}.

\boldstart{Face segmentation:} Accurate face segmentation masks are essential. We use the aligned face mesh as a cue to 
segment the input image into hair, face (including neck and ears) and background regions ($\mH$, $\mF$, and $\mB$ respectively). 
We first project the aligned face geometry onto the image plane and construct three masks -- a conservative face mask 
(strictly corresponding to face), a normal face mask (approximately face and head) and an aggressive face mask
(expansion of the normal mask), see Fig.~\ref{fig_geometry}.

Hair is typically present at the top of the head; therefore we train a GMM-based hair detector from the pixels at the
top of the normal face mask. 
The hair detector response, intersected with the normal face mask, yields a hair region.
Pixels inside the conservative face mask are assigned
the face label. Pixels outside the aggressive face mask and are not detected by the hair detector, are initially given
the background label. There might be pixels in the image that are not assigned any label. For each of these
labels, we learn a GMM-based classifier specific to that input image using color and MR8 features, and re-classify the
image. Finally, we use matting Laplacian~\cite{levinmatting} to correct the face mask.  

\boldstart{Intrinsic images:} The intrinsic image algorithm has a significant qualitative effect on our results. 
We estimate $\vsh(\vx)=\mbox{Intensity}(\vx)$ and $\rho(\vx)={\cal I}(\vx)/\vsh(\vx)$. 
Our approach (assume small dark patches are likely shadow) will result in significant errors on current evaluation 
protocols, but for the tasks we focused on, our method reliably produces better results. 
A much better albedo recovery algorithm, such as Bell et al.~\cite{bell14intrinsic} produces significantly worse makeup transfer (Fig.~\ref{belltransfer}, 
discussion below). Similar problems occur with Retinex~\cite{Land71}.

\boldstart{Face illumination:} We represent the illumination in the image using a Spherical Harmonic (SH)
model~\cite{Ramamoorthi:2001:ERI}. Per-pixel normal $\vn(\vx)$ is got by projecting the face mesh onto the image. We
then estimate the first-order SH light coefficients $\vl_s$, as:  
\begin{equation}
\vl_s = \arg \min_{\vl} \sum_{x \in \mF} (\vsh(\vx) - \renderSH(\vn(\vx), \vl))^2,
\end{equation}
where $\vsh(\vx)$ is the shading computed using our intrinsic method, $\renderSH()$ denotes the SH-based rendering of the normals $\vn(\vx)$ with the lighting coefficients $\vl$, and $x \in \mF$ restricts this optimization to the face pixels.

\boldstart{Face detail map:} Given our estimates of face geometry, albedo, and scene illumination, we can reconstruct the image as:
\begin{equation}
\vI_e(\vx) = \render(\vr(\vx), F, \vl_s) = \vr(\vx) \; \renderSH(\vn(\vx), \vl_s)
\label{eqn:rendering}
\end{equation}
Under ideal circumstances, this reconstruction will approximate the input image, i.e., $\vI_e(\vx) \approx \vI(\vx), \forall x$. 
However, the geometric model cannot represent fine-scale spatial structure on the face (for example: 
detailed folds on eyelids; small grooves and wrinkles; facial blemishes; the wrinkles on lip tissue; the folds around the outside of the lower end of the nose;  and so on).   
These details are typically the results of shading effects.
The intrinsic image algorithm we used produces an albedo map that cannot account for these details, but represents the albedo as large blotches of constant color.  As a result, 
$\vI_e(\vx)$ lacks these details. These details are represented by computing a \emph{face detail map}, $\vI_d$:
\begin{equation}
\vI_d(\vx) = \vI(\vx) - \render(\vr(\vx), F, \theta_s),
\end{equation}

%
\section{Face Editing}

As can be seen from Eqn.~\ref{eq_detail1}, our model allows us to manipulate face appearance in three ways. We can change the illumination in the scene by rendering the 
face under a new set of lights; we can edit the face appearance by editing the albedo layer; we can also edit the fine-scale details of the face by manipulating the detail map. 
The novel face image is then given by:
\begin{equation}
\vI^{(n)}(\vx) = \render(\rho^{(n)}(\vx), F, \vl^{(n)}) + \vI^{(n)}_d(\vx),
\label{eq_detail2}
\end{equation}
where $\vl^{(n)}$ represents the new illumination, $\vI^{(n)}_d(\vx)$ is a (potentially) new detail map, and $\rho^{(n)}(\vx)$ represents new albedo.

\subsection{Face Relighting}

Since we have already reconstructed the face albedo and geometry, we can render the face with different lighting effects. Our current implementation supports three distinct types of lights: SH lighting, directional lights, and colored spot lights.  SH lighting is our primary source of light, and it enables soft lighting effects~\cite{Ramamoorthi:2001:ERI}.  For other light types (i.e., directional and spotlight) we support casting shadows by leveraging the geometry stored in a depth buffer.  For colored spot lights, we separate shading into three distinct channels to allow for finer grain control of light color.

\boldstart{Rendering and compositing non-face pixels:}
While we estimate the albedo at every pixel of the image, we reconstruct geometry only on face pixels. This means that we can only render new shading inside face region, and we must extend this information to cover other regions of the image such as neck, ears and background.  We assign each pixel outside the face mask to its nearest shading value inside the face mask, and then smooth the shading.

This gives us the new shading field $\vsh^{(n)}(\vx)$ across the full image. In addition, we also have the original reconstructed shading, $\vsh(\vx)$. We blend the two shading fields
differently according to different labels. For face region, new shading  $\vsh^{(n)}(\vx)$ is used. Hair has a characteristic specular appearance, and in order to preserve it, we retains 
more of the original shading at bright pixels (which are more likely to be specular highlights). The shading for the background is a constant mixture. The final shading is smoothly blended 
from the three areas.

\boldstart{Scattering:}
There are subsurface scattering effects~\cite{JIMENEZ2009_TAP} on human faces, but this is expensive to simulate exactly. We simulate this effect by
smoothing the rendered shading field. Comparison figures both with and without smoothing is provided in the supplementary materials.

\subsection{Makeup Transfer}

We transfer makeup effects from a reference face to an input face. We want our results to meet these important qualitative criteria: 
{\bf Faithfulness:} result images should preserve the input's distinctive facial features, particularly around the eyes, nose, and lips. 
{\bf Realism:} the transferred makeup should look like real makeup, so that caking, finger smears, powder texture and so on should be preserved; 
and relit images should look like pictures taken under a real illumination field. 
{\bf Predictability:} the pattern, color and qualitative appearance of the transferred makeup should look like the reference.

To transfer colored patterns, we need to replace input image albedo $\rho^{(i)}(\vx)$ with reference albedo 
$\rho^{(r)}(\vx)$ for most regions of the face. Define $\vI_d^{b(i, r)}$ as a detail map somehow blended by input face detail map $\vI_d^{(i)}(\vx)$ and reference detail map
$\vI_d^{(r)}(\vx)$. Our transfer process is represented as
\begin{equation}
\vI^{(n)}(\vx) = \render(\rho^{(r)}(\vx), F^{(i)}, \theta^{(n)}) + \vI_d^{b(i, r)}(\vx)
\label{eq_detail2}
\end{equation}
where $\theta^{(n)}$ could be either $\theta^{(i)}$ or some new illumination (for relighting).

\begin{figure}[t]
\centering
\includegraphics[width=\textwidth]{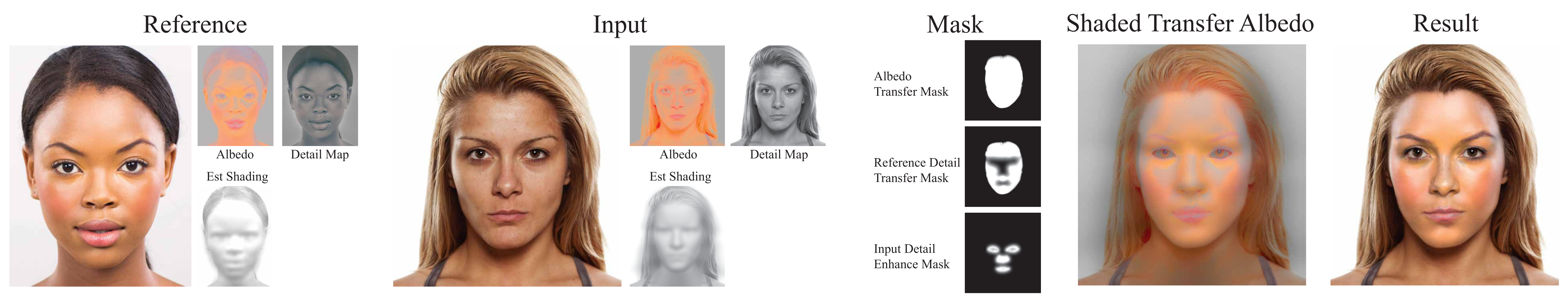}
\caption{ A makeup transfer example, which includes albedo $\rho(\vx)$ , rendered shading $\render(\rho(\vx), F,
  \theta)$, detail map $\vI_d(\vx)$, the masks used to blend detail maps, and results.
}
\label{figmaps}
\vspace{-4ex}
\end{figure}

Our algorithm proceeds as follows.  We transfer the whole albedo layer from reference to input face.  We construct a new detail map $\vI_d^{b(i,r)}$, which is a blending of reference and input detail maps.  The blending preserves the input detail map around eyes, nose and mouth, and  reference details elsewhere (Fig.~\ref{figmaps} shows our detail transfer process, and  
examples of intermediate results).   We then choose some illumination $\vl^{(n)}$ (which might be just the input illumination $\vl_s$) and use Eqn.~\ref{eq_detail2} to composite. Experiment shows $\vI_d^{b(i,r)}$ is crucial to the success of our approach. If other intrinsic image methods are used (such as Bell et al.~\cite{bell14intrinsic}), facial shadows 
etc. will mix into albedo (rather than only in detail map), and when transfer is performed, the input face appears to have lost distrinctive facial features (Fig.~\ref{belltransfer}) . 

 \boldstart{Boundary processing:}  Boundaries need to be processed more carefully for makeup transfer because portraits have a variety of face shapes, face sizes and hairlines, 
and misclassifying hair as face will create problems. We shrink the face boundary to remove
potential hair pixels, then use mirror mapping along the boundary to expand the image (we copy a 30 pixel width ring back and forth).  
Our approach is fully automatic, however, some level of user interaction could produce even better results.

%
\subsection{Face Detail Transfer}
\label{sec_detailtrans}

This section focuses on transferring face details from a reference image $\vI^{(r)}$ to an input image $\vI^{(i)}$, while preserving the general lighting of the input photograph. Different parts of the face have different material properties and fine-scale geometric variation. We account for this by transferring the detail map in nine standard components: left eyebrow, right eyebrow, left eye, right eye, forehead, nose, mouth, left cheek and right cheek. We can preserve the base face shape, and edit face details by transfering one component to all components.

In order to transfer detail maps between $\vI^{(r)}$ and $\vI^{(i)}$, we need pixel-accurate alignment between two images. We already have an initial alignment in the form of the aligned morphable face model. In some cases, there might still be some small misalignments, and we fine tune the detail map alignment with optical flow based warping~\cite{cliu-2009}. 
Makeup transfer also takes this approach to improve correspondence.

A Figure in the supplementary materials shows an example of a shiny nose (caused by specular highlights) being transferred from one source image to a variety of target faces. The alignment and warping ensure that this transfer produces reasonably realistic results.
\vspace{-3ex}
\begin{figure}[t]
\centering
\includegraphics[width=1.0 \textwidth]{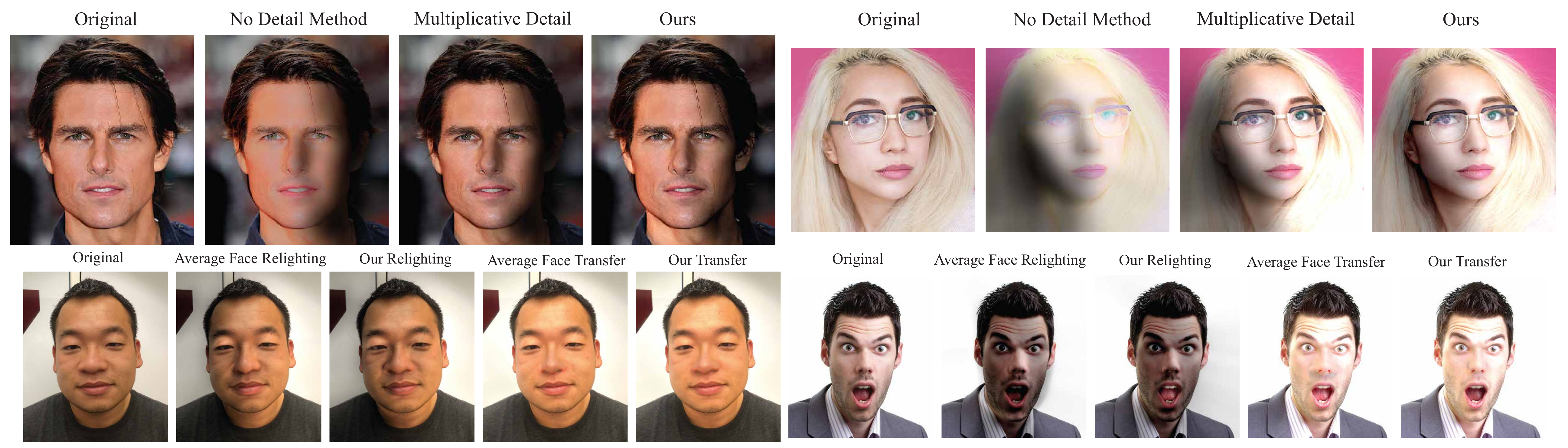}
\caption{\textbf{First Row:} Baseline rendering methods compared to ours. Removing detail maps produces overly smooth faces, and using multiplicative details loses 
detail in dark or bright areas. \textbf{Second Row:} This figure compares harsh light relighting and detail transfer results for the average face and our methods. The 
average face leads to obvious failures especially when lighting is harsh or the face has a strong expression.}
\label{fig_noVar}
\vspace{-3ex}
\end{figure}

%
\section{Results}

For all face edits, we evaluate our methods qualitatively by direct image comparisons.
Previous work has mostly evaluated qualitatively, but Wang et al. report face recognition rates for face relighting~\cite{Wang2}.  These do
not directly evaluate whether our results convince humans. We therefore evaluate quantitatively with a user study for face relighting
and detail transfer. For makeup transfer, we use an online attractiveness evaluation system to quantitatively evaluate how much we can make people look more attractive. 
All the results in this paper are best viewed at high resolution. Please refer to the supplementary materials for more examples and comparisons.

\subsection{Relighting}

We compare our method to various baselines in Fig.~\ref{fig_noVar} and in the supplementary materials. 

\boldstart{Average Face: \label{method_avg}} We register the average face mesh to the image, but all other steps in our system
remain unchanged, allowing us to evaluate the importance of our morphable face model.  This baseline performs
poorly, because not morphing the geometry leads to poor registration of nose and mouth (Fig.~\ref{fig_noVar}).

\boldstart{No Detail Map: \label{method_nodetail}} When the detail map is not applied, the results have too little spatial detail. This is due to
our approximate (and overly smooth) geometry, albedo, and shading estimates.

\boldstart{Multiplicative Detail (Shading Ratio): \label{method_ratio}}  A natural variant of our method is to compute a
multiplicative detail map, rendering the new image as
\begin{equation}
I^n (\vx) = I(x) \frac{\render(\rho(\vx), F, \theta^{(n)})}{\render(\rho(\vx), F, \theta_s)}
\end{equation}
and so allowing a comparison with a form of ratio image construction. As Fig.~\ref{fig_noVar} shows, 
this approach creates odd-looking materials and fails frequently, especially in areas where the ratio is large or small.

\begin{figure}[t]
\centering
\includegraphics[width=0.7 \textwidth]{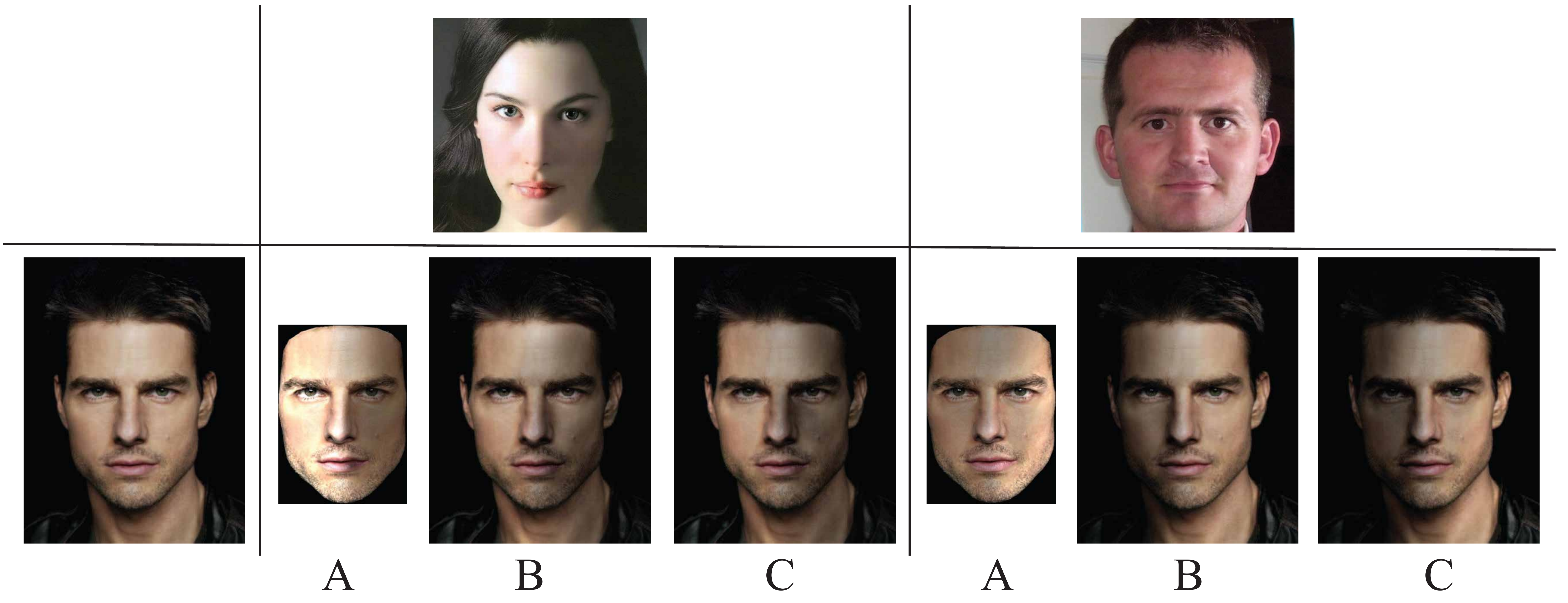}
\vspace{-2ex}
\caption{A qualitative relighting comparison between our method and Chen et al.~\protect \cite{ChenCJZ11} for one source image and two different targets. In each case, 
A comes from~\cite{ChenCJZ11} (lighting transfer). B is our result with a
manual choice of spherical harmonic light to match target lighting, and C is our method with manually choice of directional light to match target lighting.}
\label{fig_cmpLightTrans}
\vspace{-2ex}
\end{figure}

\begin{figure}[t]
\centering
\includegraphics[width=1.0 \textwidth]{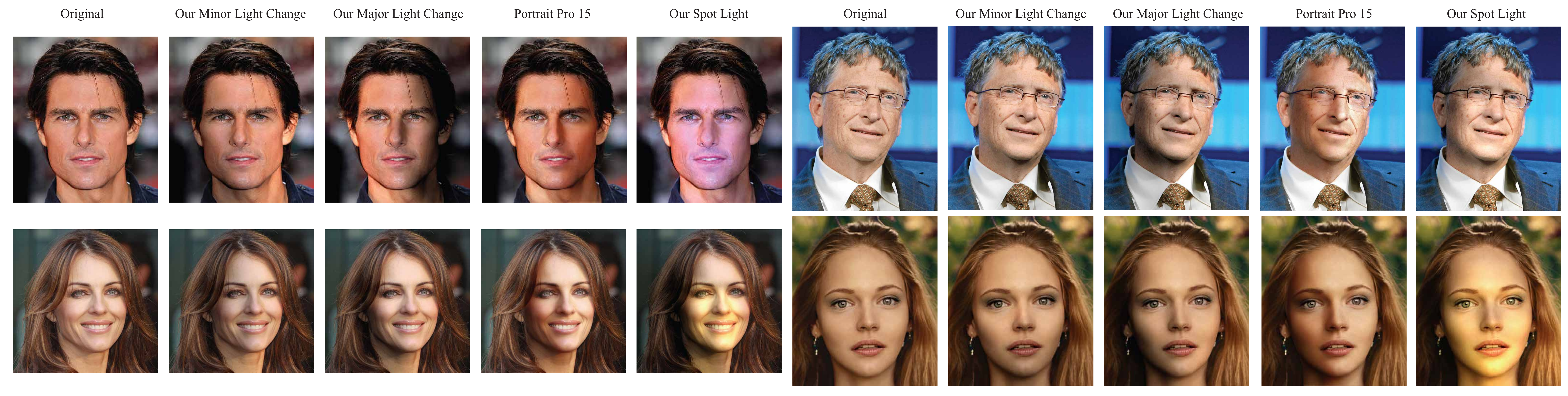}
\vspace{-5ex}
\caption{Face relighting. Each example shows the original image, our results with minor/major changes to lighting, a result from Portrait Pro 15, and our addition of spot lights.  Best viewed at high resolution in color.} 
\label{fig_relighting}
\vspace{-3ex}
\end{figure}

\boldstart{Portrait Pro 15: } It is the best available software for face relighting, but allows only minor
changes in lighting before suffering from significant artifacts.

\boldstart{Our relighting}
We have designed an interface to interactively edit spherical harmonics lights, directional lights and spot lights. We
relight images using: (a) minor changes to spherical harmonic lighting; (b) significant changes to spherical harmonic
lighting; and (c) introducing spotlights. In Fig.~\ref{fig_relighting}, we compare our results with the commercial software Portrait Pro 15. 
Even small lighting changes can cause the Portrait Pro 15 results to have noticeable artifacts; facial skin often
appears unnatural.  In contrast, even complex lighting edits produced with our method still look natural. This is because our detail map preserves 
perceived material properties of skin.  We also compare qualitatively with a recent lighting transfer
method~\cite{ChenCJZ11}, see Fig.~\ref{fig_cmpLightTrans}. Quantitative comparisons show in the user study.

\subsection{Makeup Transfer}

\begin{figure}[t]
\centering
\includegraphics[width=0.95 \textwidth]{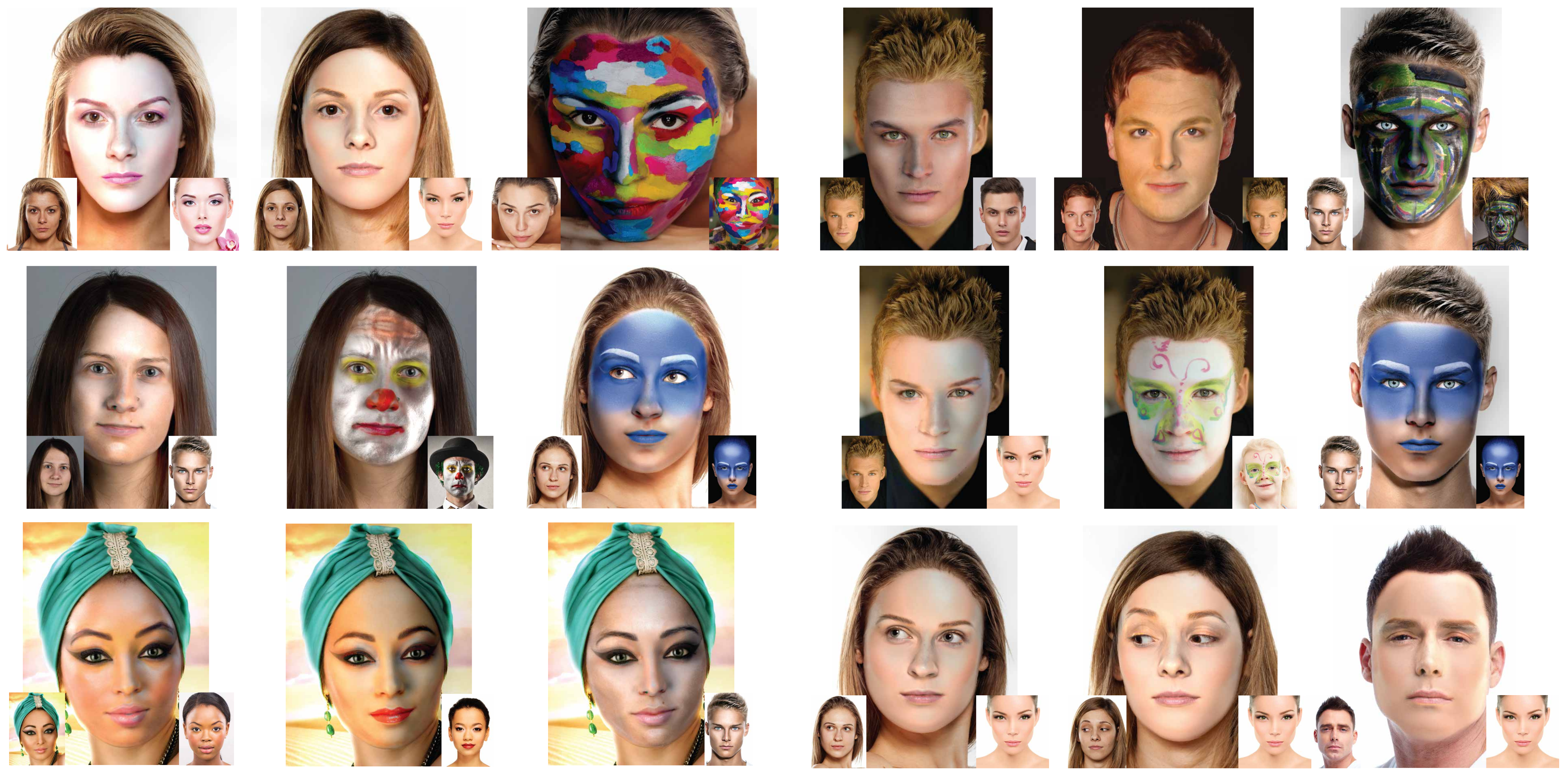}
\vspace{-3ex}
\caption{Our makeup transfer results for various situations: same gender, across gender, from multiple references to one input, 
and from one reference to multiple inputs. }
\label{fig_allMakeup}
\vspace{-5ex}
\end{figure}

Our makeup transfer is fully automatic, and works well on both normal makeup and unusual (wild) makeup. We evaluate results qualitatively by 
faithfulness (does the input face still look like themself?), realism (does the input face look like they are wearing real makeup?) and predictability 
(does the makeup on the input look like the makeup on the reference?). 
For normal makeup, we can compare with previous methods qualitatively; for wild makeup, there are no previous methods. 
We evaluate quantitatively by investigating the effect of makeup on online services that predict age and attractiveness from face images.
Fig.~\ref{fig_allMakeup} shows makeup transfer of one reference to multiple inputs (to evaluate predictability) and
from multiple references to one input (to evaluate faithfulness).  It also shows we can transfer makeup across gender.

\begin{figure}[t]
\centering
\includegraphics[width=\textwidth]{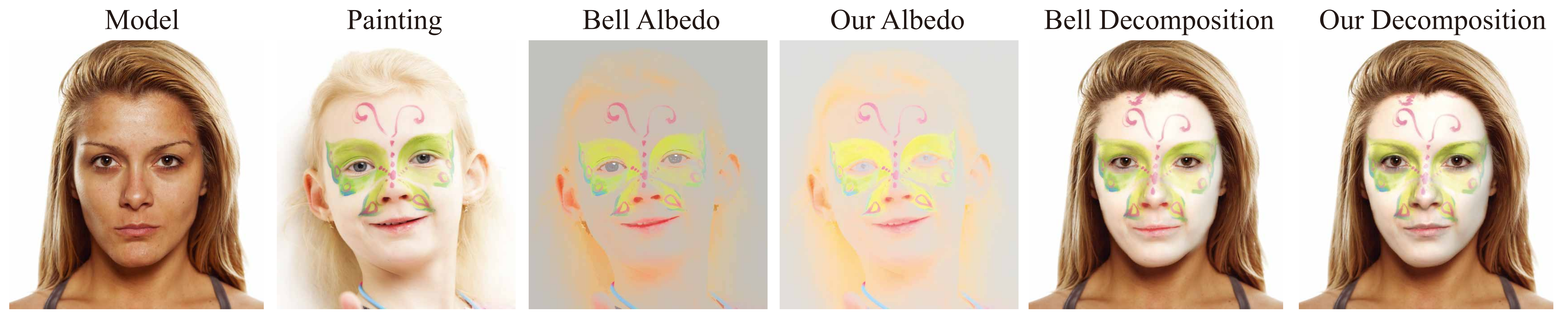}
\vspace{-5ex}
\caption{The albedo produced by Bell et al. \protect
  \cite{bell14intrinsic} method, compared with that produced by our
  method.  Their method retains some face detail in the
  albedo map (small shadows around nostril and lips); as a result, in
  the transferred image, the shape of the input's nostrils and mouth in the
  transfer ``look like'' those features in the
  reference.  The effect is further explored in the video submission.}
\label{belltransfer}
\vspace{-3ex}
\end{figure}

\boldstart{Different intrinsic images:} Fig.~\ref{belltransfer} shows the effect of using a different intrinsic image method 
(a state-of-the-art method~\cite{bell14intrinsic}).  Bell et al.'s method allocates small shadows to the albedo map, resulting in dark lips and nose 
region in both reference and input's albedo map. As a result, these shadows are missing from the detail map, so the transferred image result 
from Bell's method keeps some of the reference's facial structure (the shape of the input's mouth, the shadows in the input's nostrils, 
and the shape at the end of the input's nose). Video contains more intrinsic image comparisons. 

\begin{figure}[t]
\centering
\includegraphics[width=1.0 \textwidth]{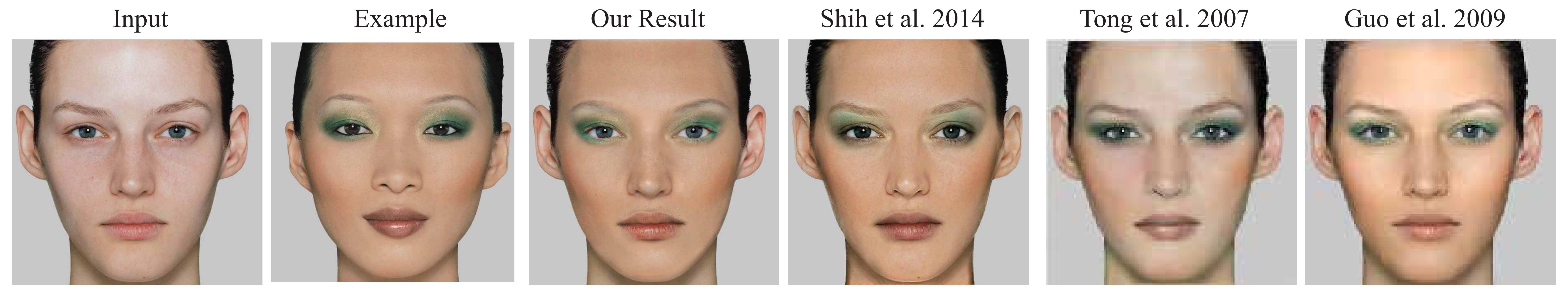}
\vspace{-5ex}
\caption{A comparison to the methods of Shih et al., Tong et al., and Guo et al. Our face 
reference points were adjusted by hand to give the best results (Shih et al. allow manual 
adjustment of reference points and masks). Results of Tong et al. and Guo et al. are far
behind and even don't get the makeup color right. The skin of Shih et al's result looks fake, especially around cheek. 
There are ringing artifacts at chin, left eyelid and nose in their result, and their eye makeup 
is lost during transfer. Our results looks more natural, eye makeup is preserved, and our method is numerically stable, so 
does not have numerical artifacts. {\em Best viewed at high resolution, in color.}
}
\label{comparison}
\vspace{-2ex}
\end{figure}

\boldstart{Comparisons:}  Fig.~\ref{comparison} shows a comparison to recent methods with normal light makeup.  
Shih et al.~\cite{Shih14:StyleTransfer} is generally successful, but loses eye makeup, and the skin around cheek is a little unnatural.
Their result also suffers ringing artifacts at nostrils, chin and around the eyes.  Earlier methods (Tong et al. \protect \cite{tong2007example} and Guo et al. \protect \cite{guo2009digital}) 
are less successful. These methods are not likely to work with stronger makeup and wild makeup, and they also don't support multiple face editing operations.
The supplementary material contains more results and further comparisons with other methods~\cite{Li15:Makeup}, etc. 

\begin{table}[h!]
\begin{center}
\resizebox{0.65\textwidth}{!}
{
\begin{tabular}{ | p{1.5cm} | p{2cm} |  p{2.5cm} | p{2cm} | p{2.5cm} |  }
  \hline
  & \multicolumn{2}{c|}{Normal Makeup} & \multicolumn{2}{c|}{Wild Makeup} \\
  \hline
  & \quad Age (year) & Attractive (1-6) & \quad Age (year) & Attractive (1-6) \\
  \hline
  Men & \quad \quad -3.8 & \quad \quad 0.6 & \quad \quad -1.6 & \quad \quad -0.2 \\
  \hline
  Women & \quad \quad -2.3 & \quad \quad 1.5 & \quad \quad -1.6 & \quad \quad 0\\
  \hline
\end{tabular}
}
\vspace{0.5ex}
\caption{Age and attractive score change after applying our makeup transfer algorithm. Age change is in years, men with normal makeup looks 3.8 years younger; 
attractiveness has discrete values range from 1 (least) to 6 (most), women with normal makeup look 1.5 levels more attractive. We care about the best improvement 
for each person using a small number of references. The number in the table is the average of the best 
improvement. Detailed table see supplementary materials.}
\vspace{-10ex}
\label{tb:howhot}
\end{center}
\end{table}

\boldstart{Makeup and attractiveness:}
Several websites predict age and attractiveness for a face image; we use http://howhot.io, which predicts age and evaluates attractiveness 
from 1 (least) to 6 (most). Table~\ref{tb:howhot} summarizes the results, and detailed table
is in supplementary materials. Transfer of naturalistic makeup tends to make a person look younger and more attractive; while wild makeup's effects depend
on images, and on average lower the attractiveness score.  Our approach quite  reliably yields a substantial, automatic improvement in attractiveness score 
by allowing an input to search a small set of references for the transfer that yields the best improvement.
We anticipate building a simple app that allows users to manipulate their photo profiles.

\subsection{Detail Transfer}

\begin{figure}[t]
\centering
\includegraphics[width=1.0 \textwidth]{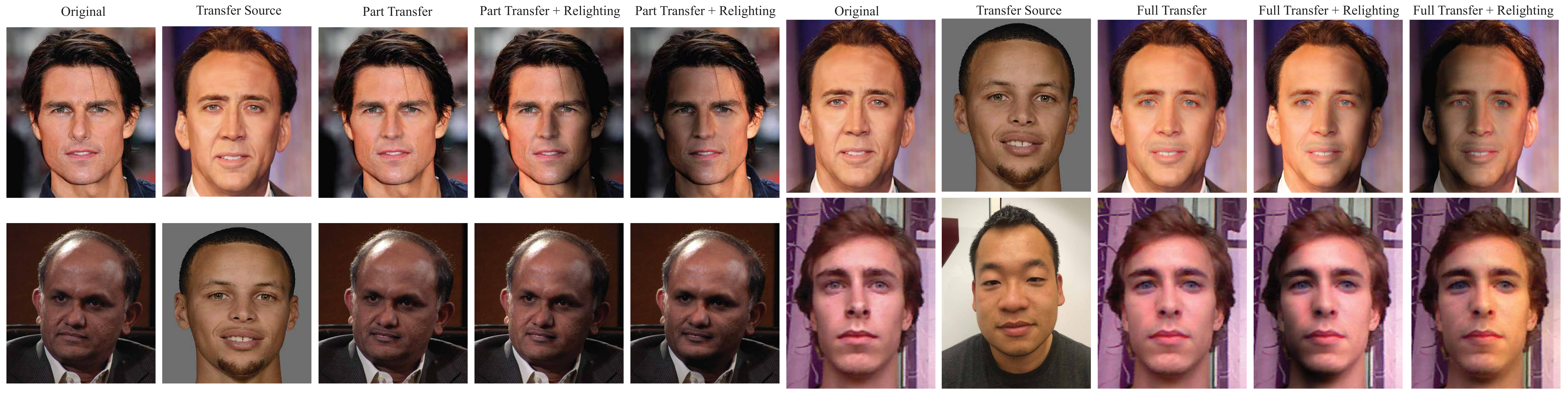}
\vspace{-5ex}
\caption{We show results of combining relighting with detail transfer either in one of nine regions or the entire face, more examples and marked transfer regions are available
in supplementary materials. }
\label{fig_partfulltransfer}
\vspace{-2ex}
\end{figure}

\begin{figure}[t]
\centering
\includegraphics[width=1.0 \textwidth]{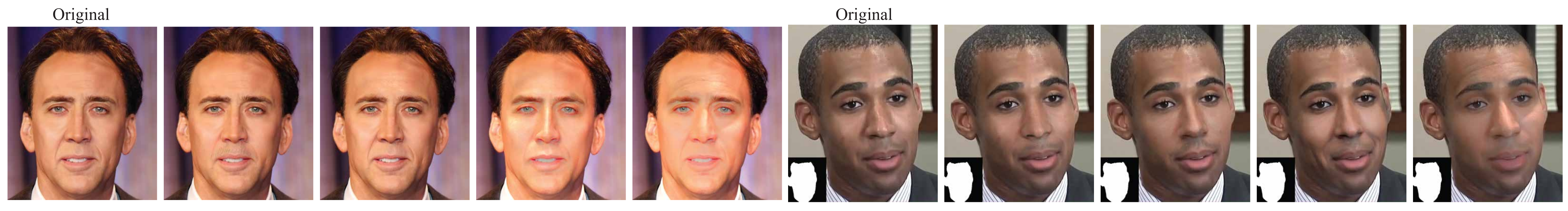}
\vspace{-5ex}
\caption{ Each example shows a series of detail map transfers using our methods, illustrating how these
transfers change the material attributes of skin.  The original image in the second example
comes from~\cite{Boyadzhiev2015}, who also edits material attributes by per image interaction (rather than
non-parametric transfer).  Comparable images appear in that paper, and are duplicated in the supplementary
materials. Supplementary materials also contain our reference images and more examples. }
\label{fig_multiToOne}
\vspace{-5ex}
\end{figure}

Our method can transfer face details like skin material attributes, wrinkles, scars, small facial expressions and even the shape of nose and mouth, while
retaining the original geometric model of the face.  Fig.~\ref{fig_partfulltransfer} shows detail transfer for areas of the face, and entire faces.
By grouping and using attributes of source images, we can reasonably control the detail transfer process, see supplementary materials. In comparison 
to~\cite{Boyadzhiev2015}, which offers users the ability to interactively edit face material attributes, our method offers a non-parametric transfer of 
a face region or whole face, see Fig.~\ref{fig_multiToOne}.
\vspace{-3ex}

\subsection{User Study}
\label{sec_userstudy}

Relighting methods do not admit comparisons with ground truth images (we do not have ground truth; the model is never
physically correct; and the key issue is whether the error is discernible to users).  We 
performed a user study to see how natural our results are. Users are presented with two images, and
must decide which of two images has not been edited.  We use the error rate in this task
to judge the method; a $50\%$ error rate means the method fools users perfectly.
To make our comparison fair, we mask the unedited image when the edited image is masked.
This likely confuses users and somewhat overestimates the effectiveness of these methods.
\begin{table}[h!]
\begin{center}
\resizebox{0.5\textwidth}{!}
{
\begin{tabular}{ | c | c | c  | c |}
  \hline
  Group & Methods & Skilled & Crowd Source \\
  \hline
   & Our Minor Light & 39.3\% & 48.9 \% \\
   & Our Major Light & 26.2\% & 44.1\% \\
  Main & Our Transfer & 46.9\% & 47.1\% \\
  & Our Light + Detail & 33.1\% & 37.6\% \\
  & Our Average  & 36.4\% & 44.4\% \\
  \hline
  & Average Face Relight & 13.8\% & 43.7\% \\
  Ablation & Average Face Transfer & 13.8\% & 8.0\% \\
  & No Detail Map&  3.5\% & 9.7\% \\
  & Multiplicative Detail  & 3.5\% & 10.5\% \\
  \hline
  & Chen 11~\cite{ChenCJZ11} & 19.0\% & 20.0\% \\
  Baseline & Wang 09~\cite{Wang2} & 22.4\% & 14.7\% \\
  & Chen 10~\cite{Chen} & 19.0\% & 10.6\% \\
  & Portrait Pro 15 & 25.9\% & 32.3\% \\
  \hline
  Objects(graphics) & Karsch~\cite{Karsch2} & \multicolumn{2}{c|}{35.5\%} \\
  \hline
  Objects(real) & Liao~\cite{Liao} & \multicolumn{2}{c|}{44.0\%} \\
  \hline
\end{tabular}
}
\vspace{0.5ex}
\caption{Error rates for our user study, higher is better, see Sec.~\ref{sec_userstudy}.}
\vspace{-10ex}
\label{tb:userstudy}
\end{center}
\end{table}

There are two groups of users. Group 1 has 29 people with some image editing background, and each finished all tasks.
Group 2 has around 100 people and is crowdsourced (CrowdFlower);  we assume these are naive users.
The error rates are listed in Table~\ref{tb:userstudy}.  In each case of:
minor change; major change; detail transfer; relight and detail transfer, we have five image pairs and we average the
error rate over image pairs and users.  For the comparison
methods (listed in Sec.~\ref{method_avg}), we have two image pairs for each task.   Tasks are presented in random
order.  For all the image pairs in the user
study, refer to our supplementary materials.   For reference, we supply error rates for object relighting methods
of~\cite{Karsch2,Liao}.   Despite reasonable concerns about uncanny valley effects, our methods can produce error rates
that match or exceed theirs.

Our study supports the conclusions: 1) users attended to tasks, because the error rate for some tasks is low; 2) both our detail transfer and
relighting methods have high error rates, meaning our method produces results people can not distinguish from natural
images;  3) it's important to use morphable models and detail maps, because the error rate drops in the absence of either
strategy; 4) the multiplicative detail map (or shading ratio) works poorly;  5) methods that mask faces solve an easier task
(because they need not relight hair, ears, neck and background) but still fool users less often than our method; 6) Portrait Pro 15 is
relatively effective, but cannot change lighting much (supplementary materials); 7) unlike the experience
of~\cite{Karsch2,Liao}, skilled users significantly outperform naive users, but can still be fooled by our method at
useful rates.

\section{Conclusion}

We propose a visual representation for face editing that combines an approximate shading model based on coarse estimates of geometry, albedo, and lighting, with a non-parametric face detail map that captures deviations like non-Lambertian reflectance, complex shading, fine-scale albedo and geometry variations. Our model can be used to solve multiple tasks: relight faces by changing the lighting parameters, transfer makeup by transferring albedo and blending detail maps, and edit face details by only transferring detail maps. 
We present a quantitative evaluation of makeup transfer by tracking the age and attractiveness from an online evaluation system. Results show that we can reasonably make people look 
younger and more attractive by our makeup transfer. 
We also present the first quantitative perceptual evaluation of face relighting and detail editing via a user study; this study shows that, to a large extent, our method produces results that users are unable to distinguish from real images. Our approach creates state of the art results on multiple face editing tasks, and in the future, we 
would like to model the face detail map more thoroughly and extend this work to more face editing applications and also applications in other fields, such as building appearance editing, etc.

\section{Appendix}
Refer to our support materials on project website for additional results and explanations. 

\begin{figure*}[h]
\centering
\includegraphics[width=0.7\textwidth]{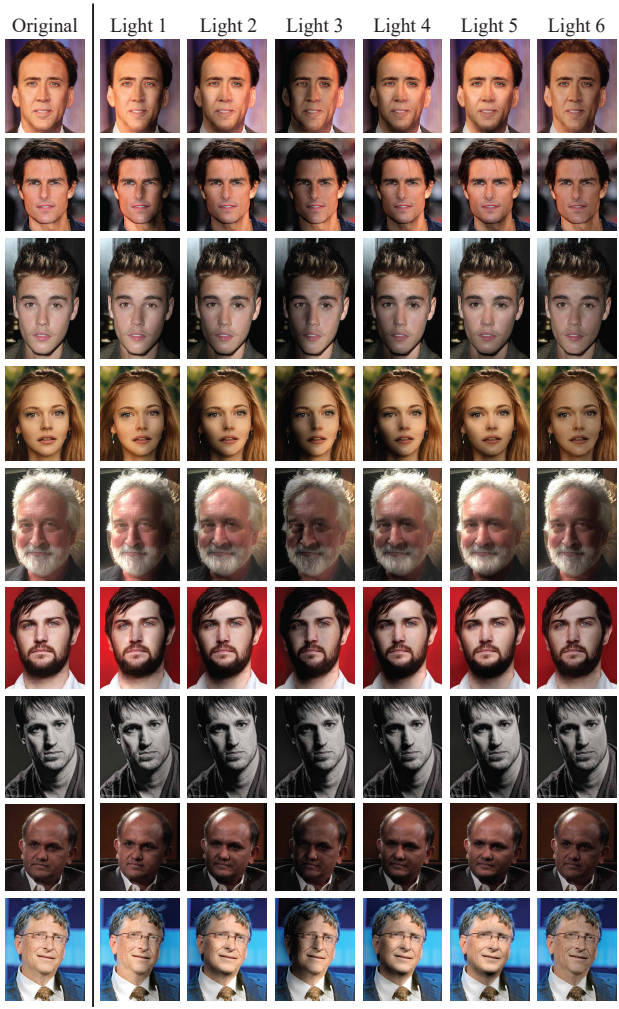}
\caption{Relighting Table. Relighting the image according to several lighting conditions. Each row is one subject, each column is one lighting condition. }
\label{fig_relightingTable}
\end{figure*}

\begin{figure*}
\center
\includegraphics[width=1.0\textwidth]{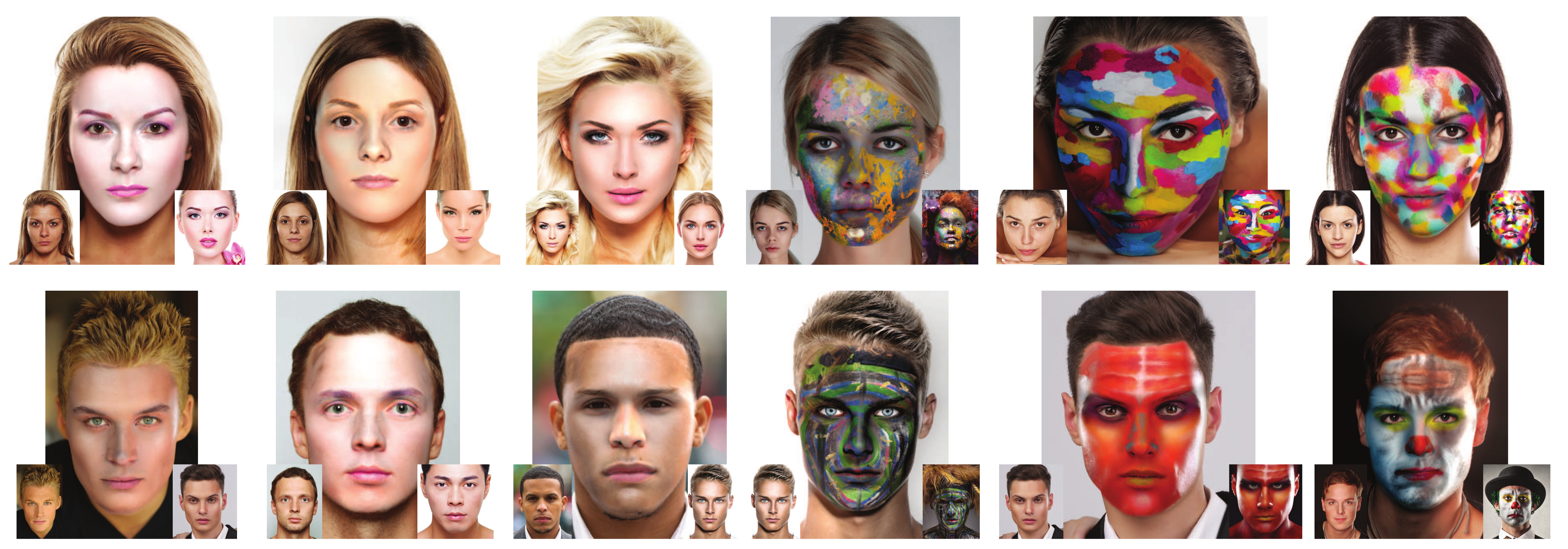}
\caption{Relighted makeup transfers from reference (lower right inset) 
image to input (lower left inset) image, for both genders
{\em Best viewed at high resolution, in color.} Further examples.
}
\label{fig:good1}
\end{figure*}

\begin{figure*}
\center
\includegraphics[width=1.0\textwidth]{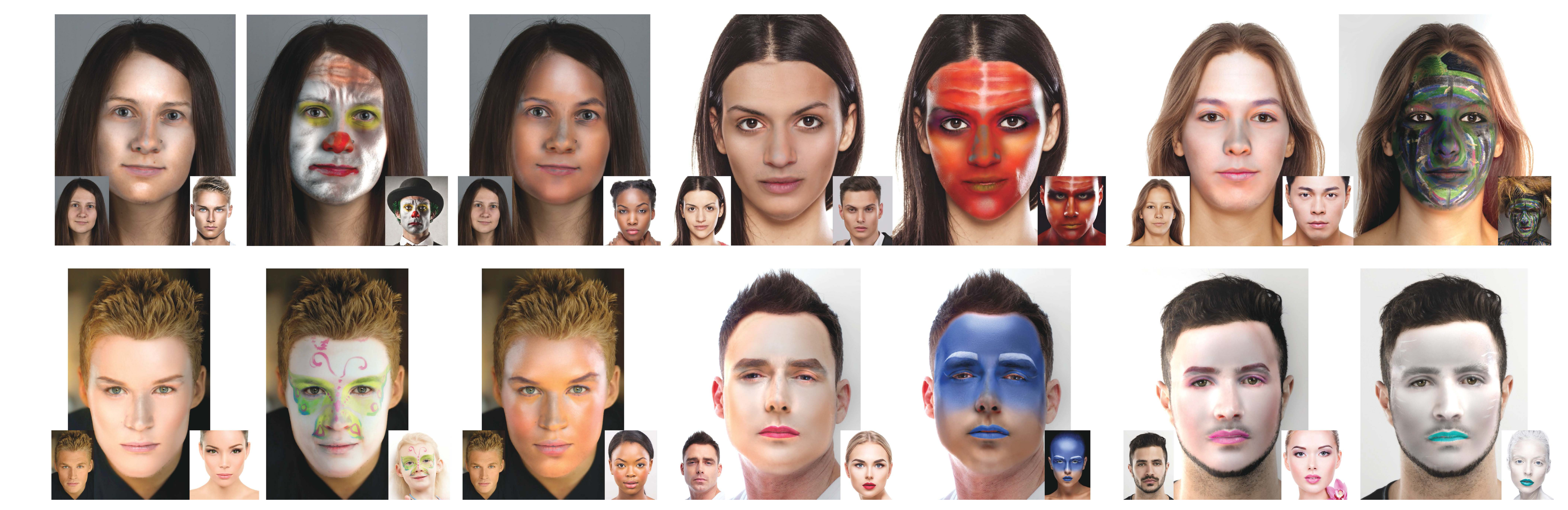}
\caption{ Across gender transfers are usually successful.
{\em Best viewed at high resolution, in color.} Further examples.
}
\label{fig:good2}
\end{figure*}

\begin{figure*}
\center
\includegraphics[width=1.0\textwidth]{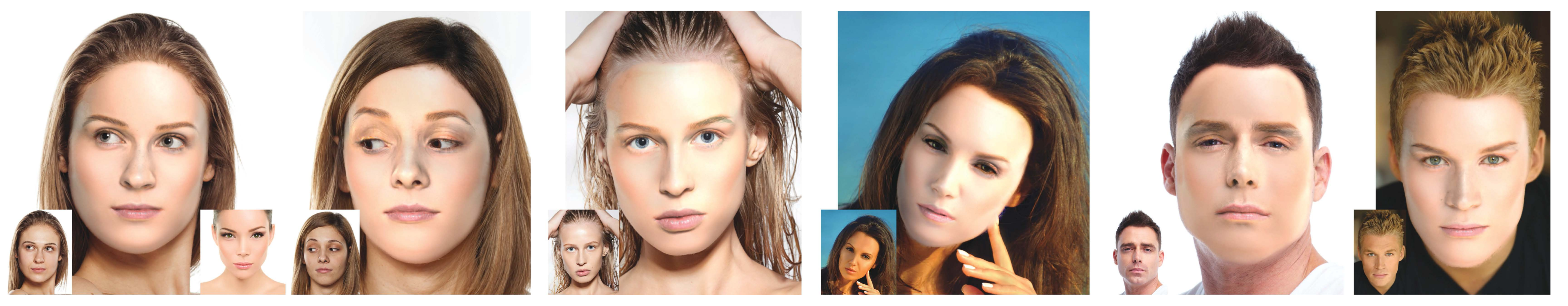}
\caption{ One reference transferred to multiple inputs. The
input retains their face shape, but appears to wear the reference's
makeup.
{\em Best viewed at high resolution, in color.} Further examples.
}
\label{fig:good4}
\end{figure*}

\begin{figure*}
\center
\includegraphics[width=1.0\textwidth]{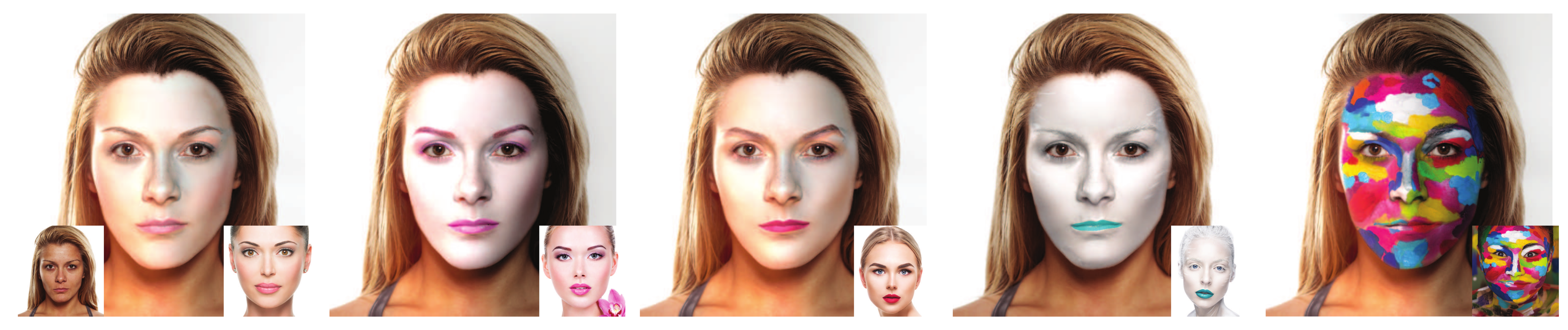}
\caption{ Multiple references transferred to one input. The
input retains their face shape, but appears to wear the reference's
makeup.
{\em Best viewed at high resolution, in color.} Further examples. }
\label{fig:good5}
\end{figure*}

\begin{figure*}[h]
\centering
\includegraphics[width=1.0\textwidth]{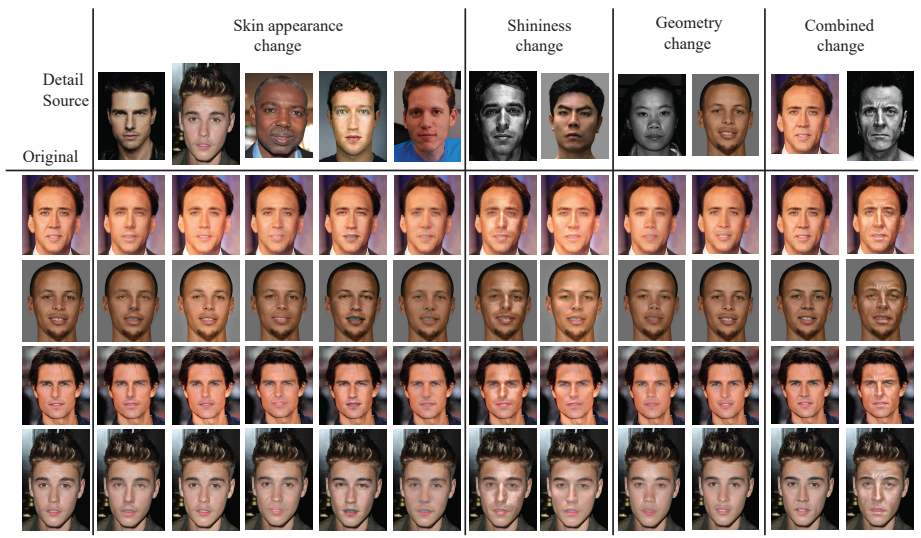}
\caption{Transfer Table. We transfer all of the nine parts of the face detail map (full transfer) from the subjects in the detail source row (first row) to the subjects in the orginal 
column (first column). Each row is one subject, and each column shares the same face detail map from the detail source person in that column. We grouped the transfer effects into
skin property change, shininess change, geometry change and combined change. }
\label{fig_fullTransferTable}
\end{figure*}

\clearpage

\bibliographystyle{splncs03}
\bibliography{egbib}
\end{document}